%% file: sn-article.tex
\documentclass[sn-mathphys,Numbered]{sn-jnl}% Math and Physical Sciences Reference Style

\usepackage{graphicx}%
\usepackage{multirow}%
\usepackage{amsmath,amssymb,amsfonts}%
\usepackage{amsthm}%
\usepackage{mathrsfs}%
\usepackage[title]{appendix}%
\usepackage{xcolor}%
\usepackage[utf8]{inputenc}
\usepackage{textcomp}%
\usepackage{manyfoot}%
\usepackage{booktabs}%
\usepackage{algorithm}%
\usepackage{algorithmicx}%
\usepackage{algpseudocode}%
\usepackage{listings}%
\usepackage{tabularx}
\usepackage{longtable}
\usepackage{caption}
\usepackage{enumitem}
\theoremstyle{thmstyleone}%
\theoremstyle{thmstyletwo}%

\theoremstyle{thmstylethree}%

\begin{document}

\title[Article Title]{Econometric vs. Causal structure learning for time-series data: Implications for policy interventions using COVID-19 data}

\author*[1]{\fnm{Bruno} \sur{Petrungaro}}\email{b.petrungaro@qmul.ac.uk}

\author[1]{\fnm{Anthony C.} \sur{Constantinou}}\email{a.constantinou@qmul.ac.uk}

\affil*[1]{\orgdiv{Bayesian Artificial Intelligence research lab, MInDS research group}, \orgname{School of Electronic Engineering and Computer Science, Queen Mary University of London}, \orgaddress{\street{Mile End Road}, \city{London}, \postcode{E1 4NS}, \country{UK}}}

\abstract{This study compares regression-based and structure-learning approaches to recovering temporal dependency structures from UK COVID-19 data and examines the limits of their use for policy analysis. We evaluate four econometric methods and eleven causal structure-learning algorithms under a common lag-expanded dataset in which edges are permitted only from past to current variables. The analysis uses 866 daily observations, with 45 variables represented by 55 numeric series after categorical encoding, and considers maximum lags of 1, 7, 14, 17 and 21 days. Four econometric methods and ten causal structure-learning algorithms successfully produced structures across all lag specifications. We compare the structure, complexity and fit of the graphs obtained under different maximum-lag specifications, and examine the stability of temporal edges across these specifications. We provide code to translate the results of these econometric methods into one of the most widely used Bayesian Network R libraries, \texttt{bnlearn}. The resulting graphs differed substantially in density, complexity, fit and stability. Agreement with the knowledge graph was limited, and low structural Hamming distance sometimes reflected graph sparsity rather than recovery of known relationships. Edge recurrence across alternative maximum-lag specifications also varied considerably between methods, indicating sensitivity to the permitted history window. Diagnostic analysis of VAR models fitted using all lagged predictors showed residual dependence, non-normality and instability, while estimation became infeasible at longer lag orders. Policy evaluation provided still stronger grounds for caution, as only a small number of mobility-to-infection links were present, few satisfied the empirical coverage criterion, and none passed that criterion under the longer-lag analyses. These findings show that imposing common temporal restrictions does not eliminate substantial methodological disagreement, and learnt temporal dependency structures should not be interpreted as policy-ready causal models without additional identification assumptions, robustness checks and adequate empirical support for the interventions being evaluated. To facilitate future work, we make the dataset and source code publicly available online.}


\maketitle

\section{Introduction}\label{sec:intro}

Focusing on associations can mislead policy design and evaluation; we need to determine whether changes in one variable cause changes in another. Traditional Machine Learning (ML) models are unsuitable for answering this question, as they focus on associations and cannot simulate the impact of hypothetical actions to inform decision-making. Answering this question requires causal assumptions that these models do not capture.

In this study, Causal Machine Learning (Causal ML) refers specifically to structure learning algorithms that learn dependency structure among available variables, with additional causal assumptions required to interpret that structure as a causal model. The two traditional classes of structure learning algorithms are score-based and constraint-based. Score-based learning searches the space of possible structures and optimises a score that reflects how well the learnt structure fits the data, whereas constraint-based learning uses conditional independence tests to infer causal structures. Hybrid algorithms that combine these two classes of learning are often viewed as a third class of structure learning.

Over the past few decades, structure learning algorithms have seen substantial methodological improvements. However, their real-world impact remains more limited than that of traditional associational ML approaches. This is because discovering causal relationships from observational data remains extremely difficult, and small errors in causal discovery can lead to large negative repercussions for causal inference. This might explain why the literature on policy recommendations remains predominantly from the field of econometrics. That is, while other fields have undoubtedly contributed to developments in optimal decision-making, economics remains one of the most deeply engaged disciplines in these topics.

\cite{bib1} is an excellent example of a widely influential discussion of methodological approaches in econometrics that favours research designs resembling Randomised Controlled Trials (RCTs). However, \cite{bib1} also mentions the developments in the ML field and their potential. Instrumental variables, regression discontinuity designs, and difference-in-differences methodologies remain the preferred tools for causal inference amongst econometricians. These methods have produced exciting work, such as \cite{bib2}, which employs a regression discontinuity design to estimate the causal effect of immigrant legalisation on the crime rate of immigrants in Italy, concluding that legalisation reduces the crime rate of legalised immigrants. However, like Causal ML, these methods have limitations. \cite{bib1} summarises issues related to weak or flawed identification strategies and problems related to this analysis's inability to generalise to other settings. 

Regression-based methods and structure-learning algorithms address different questions under different assumptions. We compare their outputs under a common temporal representation using UK COVID-19 data (see Section~\ref{sec:data}). This case study assesses how different methods recover temporal dependency structures and whether the implied intervention directions are consistent with prior substantive knowledge. Stating this as our objective implies that we will be evaluating the following:

\begin{itemize}
    \item The ability of these methods to recover plausible temporal dependency structures from observational time-series data.
    \item The stability of the recovered edges across alternative maximum-lag specifications.
    \item The extent to which the recovered structures support identifiable intervention queries under explicit causal assumptions.
    \item Whether the signs of the implied policy effects are consistent with prior substantive expectations. 
\end{itemize}

\section{Background}\label{sec:back}

The development of Causal ML for recovering cause-and-effect relationships from time-series data has been limited compared with econometrics. A few examples of current developments related to causal ML with time series data are described below: 

\begin{itemize}
    \item \cite{bib3} introduces two online causal structure learning algorithms: Online Fast Causal Inference (OFCI) and Fast Online Fast Causal Inference (FOFCI). OFCI is an online version of Fast Causal Inference (FCI) modified to handle latent variables and dynamic, time‑changing causal structures. It updates correlations as new data arrive and relearns the structure when the data indicate a significant change. FOFCI modifies OFCI to minimise learning costs and speed learning by using causal relationships learnt from previous models, while preserving model fit.
    \item \cite{bib4} introduces DOCL, a causal structure learning algorithm for real-world environments where causal structures can change unpredictably. DOCL processes streaming data in real time and adapts to changes in both the causal structure and the underlying probabilistic relationships learnt from sequential or ordered data.
    \item \cite{bib5} developed the Local Stationarity Structure Tracking (LoSST) structure learning algorithm, which assumes local stationarity in the data generation process.  LoSST dynamically tracks and adapts to changes in both the causal structure and the relationships in real time.
\end{itemize}

Although these studies are relevant, they aim to identify changes in causal structures. In contrast, we are interested in distribution shifts where the graph remains fixed but inputs, including seasonality, policy, behaviour-driven shifts, amongst others, alter observed distributions without implying new causal links.

Econometrics offers extensive time series approaches suitable for addressing this issue. Section~\ref {sec:method} provides a detailed explanation of how these methods can help recover causal structures. However, it will become clear that the methods presented work similarly to traditional causal ML methods. This is because these methods can be used to identify candidate temporal dependencies between variables while accounting for other observed predictors. The idea is similar to conventional causal discovery methods, which test for conditional independence between variables.

After learning the structures, the models can be converted into Dynamic Bayesian Networks (DBNs). A DBN is a probabilistic graphical model that represents variables and their dependencies across time. Like a standard Bayesian Network (BN), a DBN represents variables as nodes and their dependencies as directed arcs in a Directed Acyclic Graph (DAG). However, unlike a standard BN, where each variable is represented by a single node, a DBN represents each variable with multiple nodes, each corresponding to the same variable at different time points. Arcs in a DBN are drawn between variables at any $t-\ell$, for $\ell=1,\ldots,p$, and a variable at the current time $t$ whenever this variable is not independent of the past variable after accounting for all other relevant past variables. See Section~\ref{sec:var} for details on how $p$ is defined.

In terms of parameterisation, discrete BNs typically represent dependencies using Conditional Probability Tables (CPTs), while continuous BNs employ conditional probability distributions. Under assumptions concerning the correctness of the causal graph, the absence of confounding in common causes (known as causal sufficiency), and the invariance of causal mechanisms under intervention, BNs or DBNs can be given a causal interpretation as Causal Bayesian Networks (CBNs). CBNs are important to this study because they allow us to model and estimate the effects of hypothetical interventions, thereby supporting optimal decision-making under uncertainty. The intervention levels being compared must have adequate support in the relevant observed covariate histories.

Despite their utility, most of the applied work on BNs has been targeted at cross-sectional data (data collected at one specific point in time across multiple entities), or it has assumed that the data have been generated from multiple independent realisations of a process (repeated cross-sectional data pooled together and treated as a cross-sectional dataset). This is because structure learning algorithms are usually not designed to capture a system's evolution over time, the variables that compose it, or how their interrelations change over time.

\section{COVID-19 case study data}\label{sec:data}

To evaluate the usefulness of these methods in supporting policy decisions, we selected the debated policy responses to COVID-19. The COVID-19 pandemic, caused by SARS-CoV-2, posed significant global public health and policy challenges.

For example, vaccines are typically regarded as the primary strategy for managing the spread of viral diseases, and this expectation carried over to COVID-19 once effective vaccines were developed. Many viewed other actions to manage the spread of this disease as temporary measures until a vaccine became available. This may be because traditional models of infectious disease epidemics, inspired by \cite{bib6}, assume that any member of the population can infect or be infected by any other member, disregarding actual within-population interactions and how diseases spread. However, this expectation likely persisted because the economic costs of implementing social distancing measures are high.

Because modelling actual interactions within a population is challenging and social distancing measures are costly, the question is which social distancing interventions are most effective at reducing the infection rate. COVID-19 will not be the last pandemic, and since vaccines and treatments will generally not be readily available at the outset of pandemics, it is important to understand how limiting interactions within a population can more quickly control pandemics. Modelling these effects by simulating hypothetical interventions within a causal inference framework (\cite{bib7}) may help inform decision-making in future pandemics.  

We use routinely collected, publicly available, aggregated data that were widely used during the pandemic in the UK. For a detailed description of how we collected the data, please refer to \cite{bib8}. We obtained the additional relevant data for this study from the same sources cited in \cite{bib8}. These are \cite{bib9} and \cite{bib10}. The dataset used in this study contains 46 columns and 866 rows, with each column corresponding to a variable and each row to a daily outcome. The first column, 'Date', is a time indicator spanning January 30, 2020, to June 13, 2022. The dataset contains both continuous and categorical variables, as well as missing data values. Table~\ref{tab:colnames} describes the variables.

\subsection{Data missingness and imputation}

Of the 46 variables in the dataset, 40 contain at least one missing value, accounting for approximately 86.96\% of all variables. In total, 5,667 values are missing, representing 14.23\% of all data values. Table~\ref{tab:missdata} summarises missing data by variable. Key conclusions from Table~\ref{tab:missdata} include:

\begin{itemize}
    \item Vaccination data (especially the 3rd dose variables) show the highest proportion of missing values, ranging from approximately 71\% to 74\%. This pattern is expected because vaccinations were introduced later in the pandemic. However, the data itself appears delayed. UK vaccinations began towards the end of 2020, but the data starts from the end of 2021.
    \item Mobility indices (such as Apple walking mobility and Citymapper journeys) also exhibit relatively high rates of missing data (approximately 28\% and 37\%, respectively).
    \item Variables related to COVID testing and surveillance (PCR and antibody tests) have moderate missing percentages, typically ranging from 7\% to 14\%.
    \item Critical pandemic indicators such as new cases, hospital admissions, and COVID-19 deaths have relatively low missingness.
    \item The proportion of missing data increases over time, with later observations showing higher missingness rates.
\end{itemize}

To enable structure learning, we imputed the missing data.  Because the data are time series, we used a state-space model for imputation, as it accounts for temporal dynamics in estimating unobserved values. It represents time series data as a system of state and observation equations. The state equation captures the evolution of the underlying ``states" of the system over time, while the observation equation relates the states to the observed data.

The Kalman filter (\cite{bib11}) is a recursive algorithm for estimating the states in a state-space model. It is an optimal estimator when the noise is Gaussian. The Kalman filter proceeds in two steps:

\begin{itemize}
    \item Prediction: Using the state equation to predict the next state and its uncertainty.
    \item Update: Incorporating the observed data to correct the predicted state.
\end{itemize}

When data points are missing, the Kalman filter uses only the prediction step to estimate the missing value and skips the update step for that time point. Missing values were imputed separately within each series. Numeric variables were imputed in R using the na\_kalman function from the imputeTS package (\cite{bib12}), using an automatically selected ARIMA state-space model and Kalman filtering (hyperparameters: model = ``auto-arima", smooth = FALSE). For categorical variables, we created one indicator for each observed level, without an intercept, and applied the same procedure as for numeric variables. We then converted them back (from dummy variables to categorical), taking the category with the highest imputed probability as the most likely category for the missing data point.  

\subsection{Data discretisation}

We excluded the date index from learning. Except for one econometric method used in this study for structure learning, all others require numeric data. Therefore, we replaced each categorical variable with $L$ observed levels with $L-1$ binary reference-category indicators. Schools and face masks each contributed two indicators, lockdown severity and majority variant each contributed four, and season contributed three. Together with the $40$ original numeric variables, these $15$ indicators yielded $k=55$ variables for analysis.

However, we later discretised the dataset to parameterise the BNs. This choice reflects implementation restrictions in the BN software we used (\texttt{bnlearn}~\cite{bib13}), which follows the common assumption that continuous nodes may appear only as children of discrete variables. Discretisation avoids these constraints and ensures that all nodes can be treated consistently during inference. 

Because most variables represent levels of intensity, such as indices measuring transportation usage or the number of hospitalised patients, we discretised continuous variables into two categories, low and high. We used the k-means clustering algorithm for this; specifically, the 2-means variant with 50 random starts, implemented via the kmeans function (\cite{bib14}) in R. We sorted the two centroids, used their midpoint as the cut point, and labelled values at or below that point as \emph{low}, with larger values as \emph{high}. We then applied the same fitted cut point to every temporal copy of that variable, ensuring consistent state definitions across lagged and current nodes.

\section{Methodology}\label{sec:method}

This section explains how the outputs of the different methods are converted into comparable graphs, using the same variables and time lags, with edges permitted only from past to current values. We first describe the econometric methods used to learn temporal dependence structures, namely the Least Absolute Shrinkage and Selection Operator (LASSO), Least Angle Regression (LAR), James-Stein Shrinkage (JS), and Statistical Inference for Modular Networks (SIMONE). We then describe the causal ML algorithms evaluated, categorising them into constraint-based, score-based, and hybrid methods. Finally, we present the common-knowledge graph from \cite{bib8}, which we use as a benchmark for evaluating the learnt structures.

This study does not assume that all methods identify the same type of relationship. Regression methods and structure learning algorithms answer different questions under different assumptions. Regression approaches primarily capture conditional and temporal predictive dependence, whereas structure learning algorithms attempt to recover graphical structures that may support causal interpretation only under additional assumptions (see Section~\ref{sec:back}). To make the outputs comparable, we translate each method's output into the same DBN representation.

Let
\[
X_t = (X_{1t}, X_{2t}, \ldots, X_{kt})
\]
denote the vector of $k$ observed variables at day $t$. For a candidate lag order $p$, we construct the lagged data matrix
\[
\mathcal{D}_p =
\left\{
(X_{t-1}, X_{t-2}, \ldots, X_{t-p}, X_t): t=p+1,\ldots,T
\right\}.
\]
The corresponding temporal graph contains nodes for the current variables $X_{t}$ and for their lagged versions $X_{t-\ell}$, where $\ell = 1,\ldots,p$. Directed edges are allowed only from lagged variables to current variables (including a variable's own lagged version):
\[
X_{j,t-\ell} \rightarrow X_{it}, \qquad i,j \in \{1,\ldots,k\}, \ell \in \{1,\ldots,p\}.
\]
Edges from current variables to lagged variables, and contemporaneous edges among variables observed at the same time point are excluded. This restriction imposes a common temporal ordering across all algorithms.

We evaluate candidate lag specifications $p \in \{1,7,14,17,21\}$. The one-day specification serves as a parsimonious benchmark, while the longer lag windows reflect weekly reporting and plausible epidemiological or policy-response delays. These candidate lag orders are justified as a sensitivity analysis; we discuss them further in Section~\ref{sec:var}.

The econometric methods model temporal dependence using a Vector Autoregressive (VAR) process. A multivariate time series $Y_t$ follows a VAR process of order $p$ if
\begin{equation}
    Y_t = c + A_1Y_{t-1} + A_2Y_{t-2} + \cdots + A_pY_{t-p} + \epsilon_t,
\end{equation}
where $Y_t$ is the vector of time-series variables at time $t$, $c$ is a constant vector, $A_\ell$ is the coefficient matrix associated with lag $\ell$, and $\epsilon_t$ is a vector of errors. Roughly, a non-zero entry in $A_\ell$ corresponds to a directed temporal edge from a variable at time $t-\ell$ to a variable at time $t$. Thus, the VAR representation provides a
direct mapping from estimated lagged coefficients to a DBN structure. Specifics of these mappings are provided below for all econometric methods. \footnote{Please refer to \cite{bib15} for further details on VAR modelling and dynamic graphical representations.}

\subsection{Econometric Methods} \label{sec: econometric}

\subsubsection{Least Absolute Shrinkage and Selection Operator (LASSO)}

\cite{bib16} focuses on applying the LASSO for variable selection in high-dimensional graphical models. \cite{bib17} first proposed the LASSO to address modelling scenarios in which the number of predictors far exceeds the number of observations. The process in \cite{bib16} can be employed to learn the graph's structure. Focusing on a neighbourhood selection scheme using the LASSO for sparse, high-dimensional graphs, they showed that this method can recover the true graph with high probability as the sample size grows. The general idea behind using the LASSO is that real-world networks should be sparse, meaning each variable is directly influenced by only a small number of other variables. Following the original LASSO, this constrained estimation procedure tends to shrink some coefficients exactly to zero by applying an $L_1$ norm penalty to the sum of the coefficients. We select the regularisation parameter using validation data, and coefficients that remain non-zero at the selected value define the temporal dependency relationships retained in the learnt graph.

For each maximum lag and current target, we fitted a Gaussian LASSO with \texttt{glmnet} \cite{bib18}, using all eligible lagged variables, including the target's own history. We excluded constant predictors in training, and the implementation also screened for duplicated training histories. No duplicate histories were found. A constant target in training received an intercept-only model and no incoming edges. This applied to the Omicron BA.2 indicator at every lag.

We used \texttt{family = "gaussian"}, \texttt{alpha = 1}, \texttt{standardize = FALSE} (as data already standardised) and \texttt{intercept = TRUE}, requesting a regularisation path of \texttt{nlambda = 100} candidate LASSO penalty values. The minimum-to-maximum penalty ratio controls the rate of $\lambda$ values. We set it at $10^{-2}$ when the training design had more predictors than rows, and $10^{-4}$ otherwise. We controlled numerical precision with a convergence threshold of $10^{-8}$ and limited iterations to at most 100,000.

For each returned penalty, we calculated the validation mean squared error on the scaled target. Let $\lambda_{\min}$ minimise this error. The selected penalty was the largest penalty satisfying
\[
\widehat{\mathrm{MSE}}(\lambda)
\leq \widehat{\mathrm{MSE}}(\lambda_{\min})+
\widehat{\mathrm{SE}}\{\widehat{\mathrm{MSE}}(\lambda_{\min})\}.
\]
The implementation uses the sample standard deviation of the validation squared errors divided by the square root of the validation-block size. The selected model was refitted to all available rows using the same scaling and eligible predictors. We retained a lagged-to-current edge when the absolute refitted coefficient exceeded $10^{-8}$.
 
\subsubsection{Least Angle Regression (LAR)}

LAR (\cite{bib19}) aims to identify a suitable subset of predictors for linear regression, particularly when the number of predictors exceeds the number of observations. This allows us to model interdependencies in multivariate time series, where many predictors are possible. Therefore, this is useful for producing graphical models in a multivariate time series setting for the same reason the LASSO is useful. In essence, they address the same problem in a slightly different way.

The estimation procedure can be described in the following steps:

\begin{enumerate}
    \item Start with all coefficients at zero.
    \item Identify the predictor most correlated with the response.
    \item Move the coefficient of this predictor toward its least-squares value (increasing its correlation with the evolving residual).
    \item When another predictor has as much correlation with the residual as the first, proceed in a direction that is equiangular to both predictors.
    \item Continue this way, adjusting predictors based on their correlations with the residuals.
\end{enumerate}

We implement this method in R using the lars package (\cite{bib20}) with hyperparameters: \texttt{type = "lar"}, \texttt{normalize = FALSE}, \texttt{intercept = TRUE} and \texttt{use.Gram = FALSE}. The maximum number of steps in the LAR path was limited to the smaller of the number of eligible predictors and the sample size minus one.

Most of the implementation logic is the same as the LASSO, except for the validation predictions. We calculated these directly from each finite row of the fitted coefficient path, including the intercept-only step. We selected the earliest step whose validation MSE lay within one estimated standard error of the minimum validation MSE.

We then refitted the LAR path on the full sample. We extracted coefficients at the largest available finite step no later than the selected training-path step. Absolute coefficients exceeding $10^{-8}$ defined lagged-to-current edges.

\subsubsection{James-Stein Shrinkage (JS)}

\cite{bib21} introduces a simple algorithm to transform correlation networks (which are undirected) into DAGs, which can imply causal relationships. The proposed algorithm approximates the full search over all possible DAGs by assuming that, in a causal network, when two variables are correlated, once the influence of all other nodes on these two nodes is removed (partial correlation), their direct link can be directed according to a given order. The direction of the influence is determined by the time order: the lagged variable must precede the response variable in time. In time series analysis, partial correlation can help understand direct relationships between two different time series by controlling for the influence of other series. In R, this is implemented as an efficient estimator of the covariance matrix, obtained by allowing the estimated correlation coefficients to shrink toward zero and the estimated variances to approach their median. The package that enables this implementation on the lag-expanded data in R is \texttt{GeneNet} \cite{bib22}. At each maximum lag, each node was min-max scaled and \texttt{ggm.estimate.pcor} was called with \texttt{method = "static"} to obtain a shrinkage estimate of the partial-correlation matrix. The choice "static" is appropriate here because temporal copies of the variables have already been created explicitly; GeneNet's ``dynamic" option is not applied on top of them.

The estimated partial correlations were passed to \texttt{network.test.edges} with false-discovery-rate estimation enabled by default. This fits a mixture model to the partial correlations and returns, among other quantities, a posterior probability that an edge is non-zero, equal to one minus the local false discovery rate. We then used \texttt{extract.network} with \texttt{method.ggm = "prob"} and \texttt{cutoff.ggm = 0.95} to retain edges whose posterior non-null probability was at least 0.95. We retained only adjacencies between lagged and current nodes, oriented from lagged to current.

\subsubsection{Statistical Inference for Modular Networks (SIMONE)}

SIMONE (\cite{bib23}) proposes a statistical framework (and its R implementation) for inferring modules in networks. These modules, or clusters, tend to be characterised by dense connections within clusters and fewer connections between clusters.

Using a probabilistic model, it essentially captures the likelihood of connections between nodes based on their module memberships. To find the most probable modules, they proposed an optimisation procedure that incorporates regularisation to maximise the model's fit to the observed data while discouraging overly complex modules. They tested their proposed methodology on both simulated and real-world data. It effectively recovered the true modules in simulated cases but was less successful in real-world data, where it could only reveal meaningful modules.

The R implementation of SIMONE uses the following settings:

\begin{itemize}
    \item \texttt{clusters.crit = "BIC"}: This means SIMONE will pick the network structure optimising the BIC score, defined in Section~\ref{sec:evaluation}. We consider this hyperparameter choice appropriate because it is what the score-based structure-learning algorithms we describe later also optimise.
    \item \texttt{clustering = TRUE}: latent node clusters are estimated jointly with the regularised network rather than fitting a non-clustered network.
    \item \texttt{type = "steady-state"}: SIMONE is applied to the explicilty lagged data matrix. That is, the input data already contain variables of the form $X_{t-1}, X_{t-2},\ldots, X_{t-p}$ and $X_{t}$. Therefore, we use the static network setting and impose temporal ordering after estimation by retaining only edges from lagged variables to current variables. This avoids applying SIMONE's internal time-course representation to a dataset that already encodes temporal lags.
\end{itemize}

Before estimation, we min-max scaled the variables and excluded constant columns. We extracted the BIC-selected undirected network using \texttt{getNetwork(selection = "BIC")}. To obtain a temporal dependency graph comparable to the other methods, we retained only connections between lagged and current variables, orienting them from past to current observations.

\subsubsection{Model Averaging}\label{sec:mavg}

Above, we discussed four econometric methods for learning graphical structures. Structures learnt by different methods tend to be highly inconsistent. To address this issue, we also perform model averaging on the four graphs learnt by the econometric methods to obtain an overall structure. We will consider the model-averaging output in the analysis alongside the four graphs learnt independently by each econometric method. The process employed for model averaging follows the same as that employed in \cite{bib24}, and can be described as follows:

\begin{enumerate}
    \item Rank the edges according to the number of times they appear in the learnt graphs (out of four).
    \item Add the edges to the overall graph following the ranking system. 
    \item A hyperparameter is required to specify the number of times an edge must appear across the given estimated structures (four, in this case) for it to be retained. To determine this cut-off, we optimise the BIC score  (described in Section~\ref{sec:evaluation}) across the averaged structures. That is, we evaluate each possible cut-off value (i.e. retaining edges that appear in at least one, two, three, or all four structures), and select the value that maximises the BIC score. In this study, three is the number of times an edge must appear in the structures to optimise BIC.
\end{enumerate}

\subsection{Causal ML algorithms}

Structure learning is an unsupervised learning process. The traditional version is primarily based on combinatorial optimisation, where most algorithms aim to find a discrete-valued adjacency matrix. This matrix represents the presence and orientation of edges.

Each of the three learning classes described in Section~\ref{sec:intro} comes with assumptions, merits, and limitations. No consensus exists on the preferred or appropriate class. Moreover, even algorithms within the same learning class often produce widely different graphs. This phenomenon can be attributed, in part, to the fact that many algorithms operate on dissimilar assumptions about the input data and rely on distinct objective functions. Most of the considered objective functions are score-equivalent, yielding the same score for graphical structures belonging to the same Markov equivalence class. This characteristic implies that not all directed relationships can be recovered from observational data.

Before learning each causal ML structure, we min-max scaled the data. We also constructed a temporal blacklist that prohibited all arcs except those from a lagged node $X_{j,t-\ell}, \ell=1,\ldots,p$, to a current node $X_{i,t}$. The algorithms used in this study are:

\begin{itemize}
    \item PC-Stable \cite{bib25}: is a constraint-based PC-derived algorithm \cite{bib26}. It starts with a fully connected undirected graph, checks whether each pair of variables is conditionally independent, and removes the edge connecting them if so. After these tests, the algorithm attempts to orient the remaining edges.
    \item  Grow-Shrink (GS) \cite{bib27}: a constraint-based algorithm that identifies the minimal set of nodes that makes a node independent of the rest of the network to construct the network structure. This minimal set is known as a Markov Blanket. These are used to identify the neighbours (parents and children) of each node, as well as the parents of the children. This is used to incrementally identify edges representing dependencies among the variables in the data.
    \item Incremental Association Markov Blanket (IAMB) \cite{bib28}: a modified version of the GS algorithm incorporating a two-step procedure consisting of growth and pruning phases of the Markov Blanket. This pruning phase enables a more efficient identification of the Markov blanket for each variable.
    \item Fast-IAMB \cite{bib29}: an improved version of IAMB that aims to further improve efficiency by reducing the number of conditional independence tests.
    \item Interleaved IAMB (Inter-IAMB) \cite{bib28}: unlike other IAMB variants, it uses forward-stepwise selection, which improves the accuracy of the Markov blanket candidate set for each node.
    \item IAMB with False Discovery Rate (IAMB-FDR) (\cite{bib30}): incorporates FDR into the variable selection process to reduce the number of false positives.
    \item H2PC \cite{bib31}: a hybrid algorithm that first learns local structures around variables and then uses HC to learn the structure.
    \item 2-phase Restricted Maximisation (RSMAX2) \cite{bib32}: identifies a small set of candidate parent variables for each target variable based on independence tests. This reduces the space of potential graphs that have to be searched.
    \item Hill-Climbing (HC) \cite{bib33}:  a score-based algorithm that begins with an empty graph and proceeds to add, remove or reorient edges, aiming to maximise an objective function. Only a local maximum will likely be found.
    \item  Max-Min HC (MMHC) \cite{bib34}: a hybrid algorithm that starts by constraining the set of parents for each node and then applies HC to find the optimal structure in the reduced search space.
    \item Tabu search \cite{bib33}: unlike HC, Tabu, which is also score-based, sometimes makes changes that reduce the objective score, in an attempt to improve upon a local maximum solution. 
\end{itemize}

The constraint-based and hybrid algorithms require the user to specify the statistical test to be used to assess conditional independence between variables. The test selected is ``mi-g-sh". According to bnlearn (\cite{bib13}), this test is a shrinkage estimator based on the James-Stein estimator for mutual information. We left the significance level at the standard $0.05$. The score-based and hybrid algorithms require the user to input a score that assesses the quality of the learnt structure. The selected score is ``ebic-g". According to bnlearn (\cite{bib13}), "ebic-g" extends the BIC score by adding a second penalty that penalises dense networks. 

Both the test and score selected follow the principle of penalising dense networks. This seems a fair way to compare these methods with the econometric methods used in this study, which also employ shrinkage. As in previous relevant studies (\cite{bib8},\cite{bib24}), all other hyperparameters were set to their default values. The literature remains unclear on how to systematically compare results across different algorithm types with varying hyperparameters. Given the extensive number of algorithms, we invite interested readers to check the default hyperparameters in \cite{bib13}.

\subsection{Knowledge Graph}

Figure~\ref{common-knowledgegraph} below shows a visual representation of the knowledge graph assumed in this study. Although the knowledge graph is based on the previously published study \cite{bib8}, it is used here as a reference rather than a verified ground-truth causal structure. Furthermore, extending its relationships across all permitted lags is an assumption adopted in this study, rather than one established in \cite{bib8}. Some variables were grouped to simplify the visual:

\begin{itemize}
    \item The 12 mobility indexes in Table~\ref{tab:colnames} have been grouped as Mobility Indexes.
    \item The 13 variables related to tests in Table~\ref{tab:colnames} have been grouped as Tests.
\end{itemize}

For a detailed explanation of how the knowledge graph was built, please refer to \cite{bib8}.  However, some variables were not included in the knowledge graph because we used more variables than in \cite{bib8}, and the logic for integrating them was not straightforward. Specifically:

\begin{itemize}
    \item Those related to the number of vaccines administered daily.
    \item Only \emph{Tests across all 4 Pillars} was included. Further relations between capacities and tests will be for the algorithms to discover.
 \end{itemize}
 
We selected this knowledge graph (\cite{bib8}) for its relevance and its illustrative value in demonstrating the proposed methodology. \cite{bib8} clarifies the issues they faced in modelling feedback loops, and this paper builds on the static relationships defined in that study, extending them temporally. We follow the principles defined in \cite{bib8} that this dynamic process starts with the pandemic, with infection and hospitalisation rates rising before governments implement policies.

For evaluation, we expanded this base knowledge graph separately for each maximum lag $p$. Each base arc $A\rightarrow B$ was instantiated as $A_{t-\ell}\rightarrow B_t$ for every admissible lag $\ell=1,\ldots, p$. We used these lag-specific knowledge graphs, not the conceptual Figure \ref{common-knowledgegraph}, for the SHD metric defined in Section~\ref{sec:evaluation}.

\begin{figure}[ht]
    \centering
    \includegraphics[width=\textwidth,height=0.9\textheight,keepaspectratio]{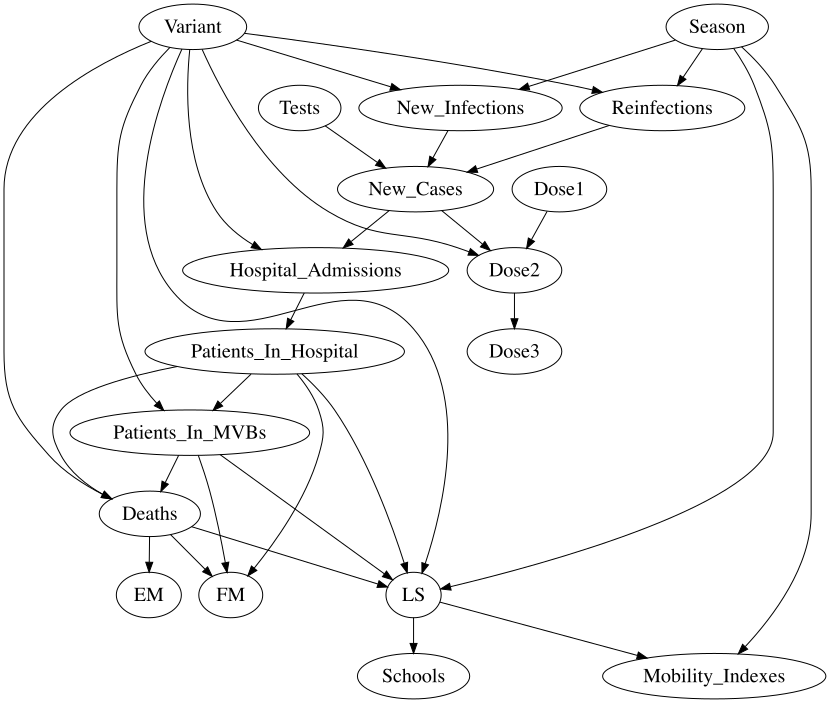}
    \caption{Conceptual knowledge graph constructed for this study based on \cite{bib8}.}
    \label{common-knowledgegraph}
\end{figure}

\section{Evaluation}\label{sec:evaluation}

After running the econometric and causal ML algorithms using their respective libraries, for each candidate maximum lag $p$, we convert the resulting DAGs into DBN models using the bnlearn library (\cite{bib13}) over the same lag-specific node set. The final graph permitted only arcs from nodes at lags $1, \ldots, p$ to current nodes. We constructed ``econometric" graphs from the selected coefficients or dependence-network adjacencies; we learnt causal ML graphs with the lagged-to-current temporal blacklist. We then analyse each algorithm's output using six distinct evaluation methods. These are:

\begin{itemize}
    \item We investigate similarities between the learnt graphs and the knowledge graph using the Structural Hamming Distance (SHD) score, which quantifies disparities between two graphs (\cite{bib34}). We also calculate a temporal-SHD, which counts the number of directed lag-specific edges that must be added or removed to make the two temporal edge sets identical. 
    \item The number of free parameters as a measure of graph complexity. For a discrete node \(X_i\) with \(r_i\) possible states and parent set \(Pa(i)\), the number of free parameters in its conditional probability table is
    \[
    d_i=(r_i-1)\prod_{X_j\in Pa(i)} r_j,
    \]
    where \(r_j\) is the number of states of parent \(X_j\). The total number of free parameters is the sum of \(d_i\) across all nodes. Consequently, model complexity can increase rapidly as the number of parents increases. We therefore also report the maximum parent-set size (maximum in-degree).
    \item The number of edges as a measure of graph density.
    \item The model selection score, BIC. We use the implementation by \cite{bib35}, defined as:
    \begin{equation}
        BIC(G,D)= \sum_{i=1}^{m} [\log Pr (X_{i}|\prod_{X_i})-\frac{|\Theta_{X_i}|}{2}\log n]
    \end{equation}
    where $G$ denotes the graph, $D$ the data, \(X_{i}\) the nodes, \(\prod_{X_i}\)
    the parents of node \(X_{i}\), \(|\Theta_{X_i}|\) the number of free parameters in the conditional probability table, and $n$ the sample size.
    \item The log-likelihood (LL) measures how well the generated graphs fit the data.
    \item We measure agreement between directed temporal edge sets using Jaccard similarity~\cite{bib36}:
    \begin{equation}
    J(E_1,E_2)=
    \frac{|E_1\cap E_2|}{|E_1\cup E_2|},
    \end{equation}
    where $E_1$ and $E_2$ denote the edge sets of the two graphs. Edges are matched by source, target and lag. Similarity ranges from zero for disjoint edge sets to one for identical edge sets.
\end{itemize}

To examine sensitivity to the maximum lag, we calculated edge recurrence separately for each successful learning method and for the consensus (model-averaged) graph described in Section~\ref{sec:mavg}. For an exact directed temporal edge $e$ at exposure lag $\ell$, recurrence was the number of successful specifications containing $e$ divided by the number of successful specifications with $p\geq\ell$. Thus, we did not penalise a long-lag edge for being impossible in shorter-lag graphs. We flagged an edge as recurrent when its recurrence was at least $0.8$ and at least two eligible specifications were available.

Because changing $p$ changes both the node set and the number of usable observations, we made direct comparisons of log-likelihoods and BIC scores within a common maximum-lag specification. Changes in parameter counts, edge counts, and SHD across maximum lags will be interpreted with caution, because the graph size and the set of possible edges also change.

Assessing the benefits and limitations of the learnt structures to support policy decisions is more complex than assessing the structure alone. One way to validate them is to compare their predicted policy effects with the observed outcomes of those policies. Evaluating COVID-19 policies is particularly challenging due to the lack of consensus on the effect sizes, but there is a consensus on the direction of the effect. For example, lockdowns are expected to reduce, rather than increase, infection rates. Under causal assumptions, the learnt structures enable us to simulate various policy scenarios and determine which interventions, particularly those aimed at reducing social interactions, are most effective in lowering infection rates. For example, if the structures consistently conclude that limiting interactions reduces virus transmission, this would support their reliability for informing future policy decisions. Specifically, we will evaluate the ability of the generated structures in the following way:

\begin{itemize}
    \item The number of causal effects between variables representing population interactions and infection outcomes. There are twelve variables related to population interactions \footnote{These are Flights (7-day moving average), OpenTable restaurant bookings (London) index, Google homeworking (Greater London) mobility index, Google workplace (Greater London) mobility index, Apple walking (London) mobility index, Google parks (Greater London) mobility index, Google retail \& recreation (Greater London) mobility index, Google grocery \& pharmacy (Greater London) mobility index, Google transit stations mobility index, TfL Tube mobility index, TfL Bus mobility index, Citymapper journeys mobility index}, and three variables related to infection rate \footnote{New cases, New infections, Reinfections}. Therefore, each structure generated by the algorithms could be used to identify up to 36 causal effects (i.e., $3 \times 12$) for each exposure lag and learnt graph.
    \item The number of times the identified causal effects match the direction the knowledge would predict; i.e., whether the effect increases or decreases. This means we will count the times when higher limits (or, equivalently, reductions) on interactions lead to reductions in infection rates.
    \item The empirical coverage of the estimated effects, assessing whether at least 95\% of the model probability mass under each intervention is supported by observed parent configurations.
\end{itemize}

\section{VAR specification and Diagnostic Tests}\label{sec:var}

This section describes how we assess the data's lag structure before estimating temporal graphs. The daily COVID-19 series may contain weekly reporting cycles, epidemiological delay windows, delayed policy responses, and slowly evolving behavioural changes. For this reason, we assess several candidate lag specifications rather than selecting a single lag order solely from one information criterion. In particular, we consider

\[
p \in \{1, 7, 14, 17, 21\},
\]

where $p=1$ acts as a parsimonious benchmark, $p=7$ captures a weekly reporting cycle, $p=14$, $p = 17$, and $p=21$ correspond to plausible epidemiological and policy-response windows.

As an initial check, we applied the \texttt{VARselect} function from the vars package in R (\cite{bib37}) with a maximum lag of 14, the feasible upper boundary. Table~\ref{tab:varselect-summary} reports the results. All information criteria (except SC/BIC) selected a lag order of $14$. The criteria used are:

\begin{itemize}
    \item Akaike Information Criterion (AIC) (\cite{bib38}) rewards goodness of fit but also penalises the number of parameters to avoid overfitting.
    \item Schwarz Criterion (SC) or Bayesian Information Criterion (BIC) (\cite{bib39}) penalises the number of parameters more heavily than AIC, often leading to more parsimonious models.
    \item Hannan-Quinn Information Criterion (HQ) (\cite{bib40}) is a compromise between AIC and BIC, with a penalty term intermediate in magnitude between the two.
    \item Final Prediction Error (FPE) (\cite{bib41}) is based on the predicted one-step-ahead prediction error. 
\end{itemize}

\begin{table}[!htbp]
\centering
\caption{Lag order selected by \texttt{VARselect} with maximum lag 14}
\label{tab:varselect-summary}
\begin{tabular}{@{}llll@{}}
\toprule
Criterion & Formula & Optimal number of lags\\
\midrule
AIC    & \(-2ln(L)+2k\) & 14 \\
SC    & \(-2ln(L)+kln(n)\) & 1 \\
HQ    & \(-2ln(L)+2klnln(n)\) & 14 \\
FPE & \(\frac{\hat{\sigma}^2_p }{n} \frac{n+p+1}{n-p-1}\)& 14 \\
\botrule
\end{tabular}
\footnotetext{where \(L\) is the likelihood of the model, \(k\) is the number of estimated parameters, \(n\) is the number of observations, and \(\hat{\sigma}^2_p\) is the estimated variance of the error term. }
\end{table}
\noindent

To assess the candidate specifications more directly, we fitted VAR models where feasible and examined residual diagnostics. Table~\ref{tab:lag-diagnostics} reports:

\begin{itemize}
    \item the AIC and BIC values from the fitted candidate models,
    \item the maximum modulus of the VAR roots, where values below one indicate a stable VAR and values at or above one suggest that some variables may be highly persistent or non-stationary, 
    \item the number of equations rejecting the Jarque-Bera (JB) test normality null hypothesis (\cite{bib42}), implemented in R via the jarque.test function of the moments package (\cite{bib43}),
    \item the number of equations rejecting the null of no serial correlation in a Breusch-Godfrey (BG) test (\cite{bib44, bib45}), implemented in R equation-by-equation via the bgtest function of the lmtest library (\cite{bib46}), and
    \item the largest absolute standardised residual Partial Autocorrelation Function (PACF)
\end{itemize}

\begin{table}[!htbp]
\centering
\caption{Diagnostics for candidate VAR lag specifications}
\label{tab:lag-diagnostics}
\small
\begin{tabular}{lccccccc}
\hline
Lag \(p\) & OLS feasible & AIC & BIC & Max root & BG rej. & JB rej. & Max PACF \\
\hline
1  & Yes & 414,265 & 428,863 & 0.9971 & 36/55 & 55/55 & 0.626 \\
7  & Yes & 394,837 & 495,453 & 0.9983 & 54/55 & 55/55 & 0.275 \\
14 & Yes & 300,138 & 501,110 & 1.0378 & 55/55 & 55/55 & 0.286 \\
17 & No  & -- & -- & -- & -- & -- & -- \\
21 & No  & -- & -- & -- & -- & -- & -- \\
\hline
\end{tabular}

\vspace{0.2cm}

\begin{minipage}{0.95\textwidth}
\footnotesize
Notes: BG rejections report the number of equations out of 55 that reject the BG null of no residual serial correlation at the 5\% level. JB rejections report the number of equations that reject the normality test at the 5\% level. The PACF bound is the approximate threshold $\pm \frac{1.96}{\sqrt{n}} =0.067$, so the reported maximum standardised residual PACF values exceed the corresponding threshold in every case. VARs with \(p=17\) and \(p=21\) are not feasible as the number of lagged regressors per equation exceeds the effective sample size.
\end{minipage}
\end{table}

The residual diagnostics indicate that none of the evaluated VAR specifications fully removes temporal dependence from the residuals. Although the maximum absolute residual PACF decreases substantially when moving from VAR(1) to VAR(7), it increases slightly under VAR(14). In all three specifications, the maximum residual PACF remains above the approximate significance bound ($\pm \frac{1.96}{\sqrt{n}} =0.067$), indicating that some temporal dependence remains unexplained. Moreover, the Breusch-Godfrey tests reject the null of no residual serial correlation in a large number of equations for every feasible specification. The Jarque-Bera tests also reject residual normality in all equations, which is unsurprising given the mixture of counts, mobility indices, policy indicators, outliers, and structural changes during the pandemic. We acknowledge these limitations, which motivates the multi-lag sensitivity analysis.

The maximum stability roots are above or very close to one in each feasible specification. The VAR(1) and VAR(7) models satisfy the stability condition, although their roots are very close to 1, whereas VAR(14) does not satisfy this condition. In a stable VAR, shocks to the system should decay over time; roots close to or above one suggest that shocks decay very slowly or may not decay at all. This result is consistent with the highly persistent nature of several variables in the COVID-19 data, including cumulative vaccination measures, policy indicators, hospitalisation series, testing variables, and mobility indices. It may also reflect non-stationarity, structural breaks, or regime changes over the pandemic period.

We estimate and compare temporal structures across multiple lag specifications. We interpret edges that appear only under $p=1$ as fragile, while edges that recur across weekly and epidemiologically motivated lags reflect more robust temporal dependencies. For higher lag windows such as $p=21$, unrestricted OLS VAR estimation is not feasible. However, our methodologies are regularised structure-learning procedures. Therefore, we proceed with the analysis of these higher lags.

\section{Results}

The first subsection focuses on structure learning and compares the learnt structures with the knowledge graph, along with metrics that describe their fit and complexity. The subsequent subsection discusses policy evaluation, focusing on assessing the quality of the learnt structures with respect to their application in decision-making.

\subsection{Graphical and modelling metrics}

Table~\ref{tab:results} shows substantial differences in the number of selected edges. At $p=1$, LAR and LASSO select 511 and 463 edges, respectively, compared with 170 for HC, 171 for Tabu, 161 for JS, and 64 for SIMONE. HC and Tabu produce the largest graphs among the structure learning algorithms (causal ML) at every maximum lag, but LASSO and LAR exceed them throughout, and JS does so for $p\geq7$.

The methods respond differently to expansions of the history window. JS grows from 161 edges at $p=1$ to 3,129 at $p=21$, while LASSO increases more moderately from 463 to 645. LAR reaches 1,629 edges at $p=17$ before falling to 1,455 at $p=21$, a 12\% reduction. SIMONE selects 64, 105 and 116 edges at $p=1$, $p=7$ and $p=14$, respectively, but no temporal edges at $p=17$ or $p=21$. Therefore, allowing additional history can remove or replace previously selected relationships. Changes in the conditioning variables, tuning and usable sample may all contribute to this pattern.

In contrast, causal-ML graph sizes change relatively little after $p=7$. HC contains 166 edges at both endpoints, while Tabu changes from 166 to 172. PC-Stable consistently selects more edges than the other constraint-based methods and approaches the sizes of HC and Tabu graphs at longer maximum lags. The BIC-selected econometric consensus retains 72, 125, 135, 137 and 138 edges across the five maximum-lag specifications, respectively, using the same three-vote threshold throughout.

\input{edge_counts.tex}

At the selected maximum lag $p=1$, the comparison with the knowledge graph initially suggests that some methods perform better than others. As shown in Table~\ref{tab:selected-metrics}, Fast-IAMB achieves the lowest SHD (239), closely followed by MMHC (240). A similar pattern emerges when considering temporal SHD, which compares the exact directed edges: Fast-IAMB again performs best (238), followed by MMHC (239). However, these results become less encouraging when we examine how many relationships the methods actually recover from the 180-edge knowledge graph. Fast-IAMB identifies only two of these edges. In contrast, LASSO recovers 50 reference edges, considerably more than Fast-IAMB, despite having a much higher temporal SHD (543). This apparent contradiction arises because SHD penalises both missing and additional edges. Consequently, methods that produce very sparse graphs can achieve relatively low distances simply by avoiding additional edges, even when they fail to recover most of the relationships in the knowledge graph.

\input{selected_metrics.tex}

\begin{figure}[htbp]
\centering
\includegraphics[width=\textwidth]{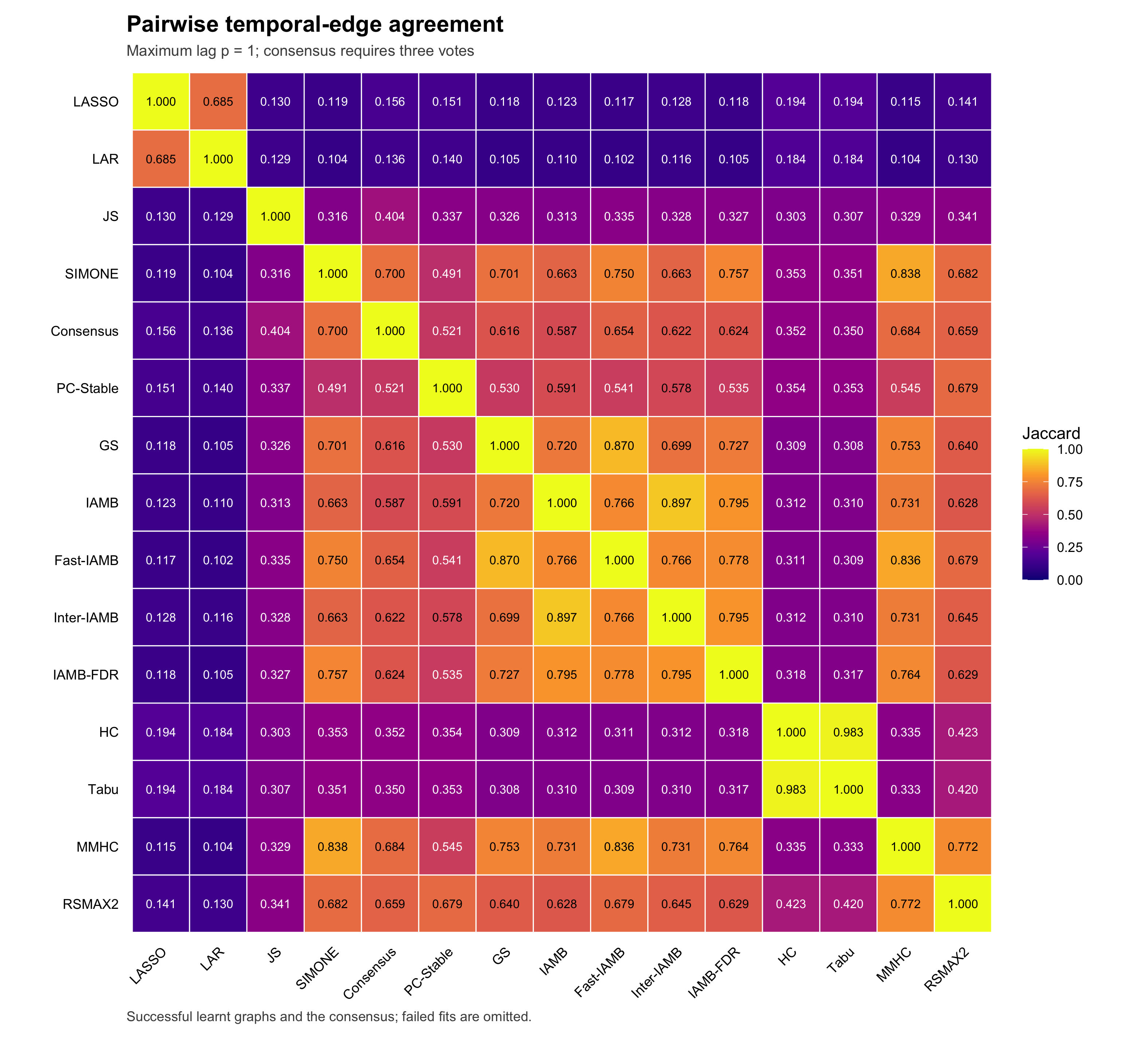}
\caption{Pairwise Jaccard similarity at $p=1$. Successful learnt graphs and the three-vote consensus are shown; unavailable outputs are omitted. A value of one denotes identical edge sets. This graphic was made using \cite{bib47}.}\label{fig:pairwise}
\end{figure}

Figure~\ref{fig:pairwise} reveals much stronger agreement between some learnt graphs. HC and Tabu are almost identical, with Jaccard similarity of 0.983. IAMB and Inter-IAMB also have significant similarities. SIMONE and MMHC have a similarity of 0.838, whereas LASSO and LAR overlap by 0.685. These results show that similarly sized graphs can have different degrees of structural agreement. Strong agreement is particularly apparent between some methodologically related algorithms, although interestingly it is not confined to such pairs. This is an interesting finding to analyse the mechanics of these algorithms in future work.

At $p=1$, a threshold of at least three votes out of four econometric methods maximises the consensus BIC among the evaluated thresholds. The resulting consensus retains 72 temporal edges: 49 are selected by all four econometric methods and 23 by exactly three. The threshold selected at $p=1$ is then held fixed across the remaining maximum-lag specifications to enable consistent comparisons. The consensus therefore identifies a relatively small common subset of the relationships selected by the individual methods.

Graph complexity differs much more sharply than edge counts alone suggest. LASSO and LAR have maximum parent-set sizes of 52 and 54, producing approximately $7.32\times10^{15}$ and $2.83\times10^{16}$ free parameters. By comparison, the consensus has at most four parents per node and 230 free parameters; JS has at most five parents and 637 free parameters, while SIMONE has at most three parents and 191 free parameters. HC and Tabu each have a maximum parent-set size of 14 and about 18,200 free parameters.

RSMAX2 has the best BIC score at every maximum-lag specification ($p=1$, $-31{,}020$), followed by PC-Stable ($-31{,}163$) and Inter-IAMB ($-31{,}597$) at $p =1$. The consensus ranks tenth among the 15 structures, with a score of $-32{,}525$. Its smaller dimension does not compensate for its poorer fit. Conversely, LASSO and LAR have relatively high log-likelihoods but receive extremely large complexity penalties. HC and Tabu are also penalised heavily because some nodes have many parents.

Appendix~\ref{app:stability} reports recurrence under the rule defined in Section~\ref{sec:evaluation}. The three-vote consensus has 90 recurrent edges out of 242 (37.2\%), compared with 22.6\% for LASSO, 20.4\% for LAR and 31.5\% for JS. SIMONE has no recurrent edges among its 139 distinct selected edges. HC and Tabu yield approximately 40\%, while higher proportions occur for MMHC (77.2\%), IAMB-FDR (68.6\%) and RSMAX2 (63.8\%). Recurrent consensus edges predominantly represent shorter delays: 78 of 90 have exposure lags of seven days or less, including 37 lag-1 edges present in all five specifications.

These proportions depend on graph size and edge distribution across lags. Lag-1 edges must appear in at least four of five specifications, whereas edges at lags 7 to 17 must appear in every eligible specification. Edges at lag 21 cannot qualify because only one specification is eligible. Edges are compared using their exact time delays, meaning that relationships between the same variables at different lags are treated as distinct edges. Recurrence therefore describes repeated time-delayed selection across the different maximum history windows.

\subsection{Policy Evaluation}

Simulating interventions in CBNs is equivalent to manipulating a variable in the model. In this study, this means, for example, a policy intervention aimed at reducing one mobility index and estimating its impact on future COVID-19 cases.  We use Pearl's \cite{bib7} do-operator to simulate interventions for causal effect estimation, where \(P(Y|do(X=x))\) represents the probability of observing \(Y\) if we manipulate \(X\) and set it to the state \(x\). 

For each learnt graph, we calculated how the predicted probability of a high outcome changes when we set the earlier mobility exposure to low instead of high. These probabilities used a uniform-Dirichlet posterior estimator with imaginary sample size 10, matching \texttt{bn.fit(method = "bayes", iss = 10)} in \texttt{bnlearn} \cite{bib13}. 

We calculated probabilities using observed combinations of predictor values and added the contribution of unobserved combinations using the prior's default probabilities. This avoided storing large probability tables or using Monte Carlo simulations. The results are point estimates conditional on the fitted model.

Since the data is discretised into two different states for continuous variables, we compute the Average Causal Effect (ACE) as follows:

\begin{equation}\label{eq:ace}
\Delta_{\ell}=P_G\{Y_t=\mathrm{high}\mid\mathrm{do}(X_{t-\ell}=\mathrm{low})\}
-P_G\{Y_t=\mathrm{high}\mid\mathrm{do}(X_{t-\ell}=\mathrm{high})\}.
\end{equation}

A negative $\Delta_{\ell}$ means a lower probability of a high outcome under the low state of the mobility indices. Here, \(X\) represents variables related to population interactions, and \(Y\) represents variables associated with the infection rate defined in Section~\ref{sec:evaluation}.

At model maximum lag $p=1$ and exposure lag $\ell=1$, the fifteen graphs yield 540 computable
$\Delta_{\ell}$ as shown in Table~\ref{tab:policy-counts}. 27 contain a direct exposure-to-outcome link and 513 (95\%) are zeros under the proposed structures. These zero effects are the result of the absence of directed paths between the relevant variables in the learnt graphs. Direct links occur only in LASSO, LAR, JS, HC and Tabu.

\input{policy_counts.tex}

Eight of the 27 direct exposure-to-outcome links (29.6\%) pass the coverage diagnostic, while the remaining 19 have low coverage. A $\Delta_{\ell}$ passes when at least 95\% falls on observed combinations under both interventions. The passing $\Delta_{\ell}$ comprise four from HC and the same four from Tabu, as shown in Table~\ref{tab:covered-contrasts}. Each has mass 1 under both intervention states.

\input{covered_contrasts.tex}

In these HC and Tabu models, setting homeworking to its low rather than high state increases the fitted probability of high new cases by approximately 1.22 percentage points and high new infections by 1.71 percentage points. Setting grocery/pharmacy mobility to its low state instead gives small decreases of approximately 0.044 and 0.140 percentage points, respectively. These results suggest that, under these models, preventing homeworking from falling provides the most effective way to reduce infections. Appendix~\ref{app:all-contrasts} reports all direct exposure-to-outcome links.

These findings are broadly consistent with evidence that SARS-CoV-2 transmission risk depends on the duration and proximity of interpersonal contacts. \cite{bib48} analyses approximately seven million contacts recorded by the NHS COVID-19 app and found that transmission probability increased with exposure duration and was associated with contact proximity. Such findings provide a plausible epidemiological mechanism through which changes in homeworking may influence infection outcomes.

\begin{figure}[htbp]
\centering
\includegraphics[width=\textwidth]{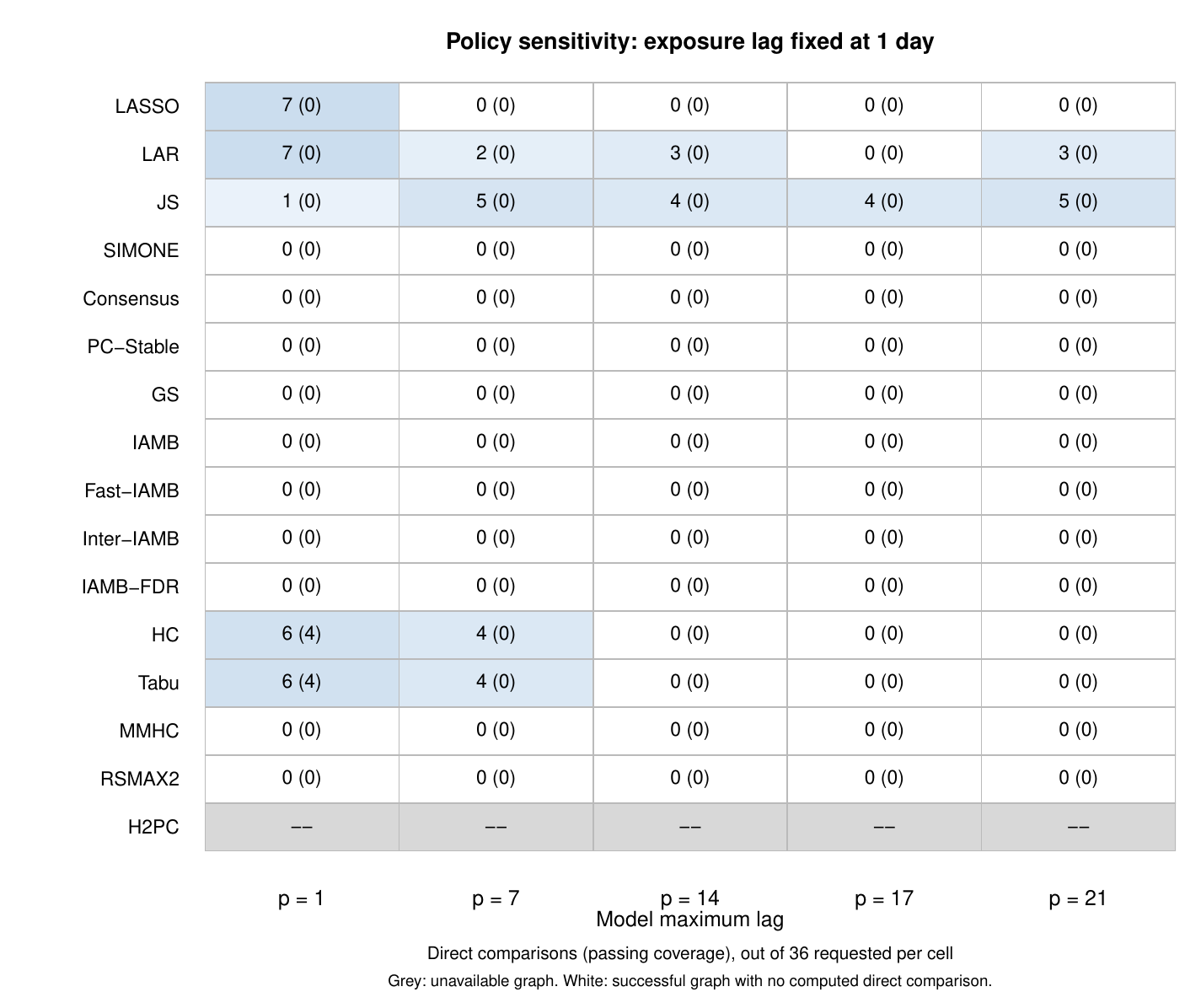}
\caption{Sensitivity of lag-1 exposure-to-outcome links to the model maximum lag, with exposure lag fixed at one day. Each cell reports the number of links, with
the number passing the coverage diagnostic in parentheses, out of 36 requested comparisons.}\label{fig:policy-sensitivity}
\end{figure}

Figure~\ref{fig:policy-sensitivity} holds the exposure lag at one day while varying the model maximum lag. Across $p\in\{1,7,14,17,21\}$, the numbers of exposure-to-outcome links are 27, 15, 7, 4 and 8, respectively, out of 540 available at each specification. These immediate lag-1 links therefore depend on the permitted history window, declining and then partially recovering at $p=21$. No lag-1 exposure-to-outcome links at a longer model maximum pass the coverage diagnostic.

We also held the model maximum lag fixed at $p=14$ and evaluated exposure lags $\ell=1,\ldots,14$. Extending the analysis to every permitted exposure lag yields no exposure-to-outcome links meeting the requirement that at least 95\% of model mass fall on observed
parent configurations under both intervention states.

\section{Conclusion}\label{sec13}

This study compared econometric and structure-learning methods using UK COVID-19 time-series data, focusing on their ability to support policy interventions. The findings reveal substantial differences between the structures learnt by different methods and highlight important limitations in their use for causal inference. These results have broader implications for policy decision-making, emphasising the need to assess the reliability of learnt structures across different methods before using them to evaluate interventions.

The results show substantial differences between methods even after imposing the same temporal restrictions. LASSO and LAR generally produced much larger graphs than the structure learning algorithms, while JS became increasingly dense as the maximum lag increased. More importantly, edge counts alone understated the differences in model complexity. At $p=1$, LASSO and LAR produced maximum parent sets of 52 and 54 variables and implied approximately $7.32\times10^{15}$ and $2.83\times10^{16}$ free parameters, respectively. In contrast, the three-vote econometric consensus contained at most four parents per node and required only 230 free parameters. RSMAX2 achieved the best BIC among the compared structures at every maximum-lag specification, while LASSO and LAR obtained relatively high likelihoods only at the cost of extremely large complexity penalties.

At $p=1$, Fast-IAMB and MMHC achieved some of the lowest SHD scores, but their exact overlap with the reference edges was small. LASSO recovered substantially more reference edges but produced a larger structural distance because its graph was much denser. Consequently, a low SHD should not be interpreted on its own as evidence that a method has successfully recovered the substantive dependency structure. Pairwise comparisons similarly showed that the strongest agreement tended to occur between methodologically related algorithms, such as HC and Tabu and between LASSO and LAR. Interestingly, there were exceptions to this point that deserve further exploration.

The multi-lag analysis further demonstrates that conclusions depend on the permitted history window. The consensus structure had 90 recurrent edges out of 242 distinct selected edges, while recurrence proportions differed markedly across individual methods. Most recurrent consensus relationships occurred at relatively short exposure lags. Some structure-learning methods, including MMHC, IAMB-FDR and RSMAX2, showed higher recurrence proportions, although these figures also depend on graph size and on which lagged edges are eligible to recur. SIMONE had no recurrent edges.

Residual serial dependence remained substantial, residual normality was rejected throughout, and the maximum stability root was close to or above one. The $p=14$ specification was unstable, while estimation was infeasible for $p=17$ and $p=21$ because the number of lagged regressors exceeded the effective sample size when using a non-regularised VAR for diagnostics. These results support sparse and regularised methods for learning high-dimensional temporal structures.

The policy analysis provides the strongest limitation on causal interpretation. At $p=1$, the fifteen available graphs generated 540 mobility-to-infection comparisons, but only 27 corresponded to direct exposure-to-outcome links. Of these, only eight satisfied the requirement that at least 95\% of the model mass under both intervention states be supported by observed parent configurations. These eight estimates came from HC and Tabu and represented the same four contrasts in each graph. When we increased the model maximum lag, the number of immediate mobility-to-infection links declined substantially, and none of the links identified under the longer-lag specifications passed the coverage diagnostic. Likewise, when we considered all exposure lags from one to fourteen days within the $p=14$ models, no queried link met the coverage requirement.

Accordingly, the main contribution of this comparison is not evidence that one family of methods consistently recovers a policy-ready causal model. Instead, the results show how strongly learnt temporal structures can depend on the learning procedure, complexity penalty, and permitted lag history, and how apparently identifiable intervention queries may lack empirical support once positivity and observed parent configurations are examined. Sparse graphs can perform well on model-selection or structural-distance criteria without recovering many substantively expected relationships, while denser graphs can fit the data more closely at the cost of extreme parameterisation.

For policy applications, temporal precedence, sparsity, stability and empirical intervention support should therefore be assessed jointly rather than relying on a single goodness-of-fit metric. Future work should investigate structure-learning procedures that more directly accommodate non-stationarity, structural breaks, latent confounding and changing pandemic regimes, as well as model-averaging approaches that incorporate edge stability and intervention support rather than fit alone.
\backmatter

\section*{Declarations}

\subsection{Data availability and access}

Code, data, replication materials, and additional documentation are available in the accompanying GitHub repository:

\url{https://github.com/br1pa/econometric-vs-causal-time-series}

\subsection{Competing interests}

The authors declare that they have no known competing financial interests or personal relationships that could have appeared to influence the work reported in this paper.

\subsection{Ethical and informed consent for data used}

This article does not contain any studies with human participants or animals performed by any of the authors.

\subsection{Contributions}

Bruno Petrungaro: Conceptualisation, Methodology, Software, Analysis, Writing-original draft preparation, review and editing. 
Anthony C. Constantinou: Writing - Review \& Editing, Supervision.

\clearpage
\begin{appendices}
\section{Variables and missingness}\label{app:data}
The tables below describe the variables and report missingness before imputation. The date index is not a graph node.
\input{variables.tex}
\input{missingness.tex}

\clearpage
\section{Edge recurrence}\label{app:stability}

Table~\ref{tab:stability-counts} reports the number of distinct selected edges and the number meeting the recurrence flag (recurrence $\ge 0.8$, at least two eligible specifications available).

\input{stability_counts.tex}

\clearpage
\section{All ACEs}\label{app:all-contrasts}

Table~\ref{tab:all-policy-contrasts} reports all 27 $\Delta_{\ell}$ at $p=1$ and $\ell=1$. The remaining 513 computed $\Delta_{\ell}$ are zero.

\input{all_direct_contrasts.tex}

\end{appendices}

\bibliography{sn-bibliography}

\end{document}

%% file: edge_counts.tex
\begin{table}[htbp]
\centering
\caption{Edge count by method and model maximum lag.}\label{tab:results}
\small
\setlength{\tabcolsep}{3pt}
\begin{tabular}{@{}lrrrrr@{}}
\toprule
Method & $p=1$ & $p=7$ & $p=14$ & $p=17$ & $p=21$ \\
\midrule
LASSO & 463 & 482 & 514 & 556 & 645 \\
LAR & 511 & 918 & 1,502 & 1,629 & 1,455 \\
JS & 161 & 1,391 & 2,214 & 2,486 & 3,129 \\
SIMONE & 64 & 105 & 116 & 0 & 0 \\
Consensus & 72 & 125 & 135 & 137 & 138 \\
PC-Stable & 109 & 155 & 157 & 156 & 161 \\
GS & 67 & 97 & 104 & 101 & 99 \\
IAMB & 74 & 96 & 100 & 94 & 93 \\
Fast-IAMB & 62 & 93 & 103 & 95 & 93 \\
Inter-IAMB & 74 & 102 & 101 & 98 & 93 \\
IAMB-FDR & 66 & 74 & 85 & 81 & 78 \\
HC & 170 & 166 & 163 & 169 & 166 \\
Tabu & 171 & 166 & 163 & 170 & 172 \\
MMHC & 61 & 79 & 74 & 75 & 73 \\
RSMAX2 & 79 & 100 & 95 & 95 & 96 \\
H2PC & -- & -- & -- & -- & -- \\
\bottomrule
\end{tabular}
\par\smallskip
\begin{minipage}{\linewidth}\footnotesize\raggedright
A dash denotes an unavailable output. The consensus requires at least three votes from the four available econometric methods: LASSO, LAR, JS and SIMONE.
\end{minipage}
\end{table}

%% file: selected_metrics.tex
\begin{table}[htbp]
\centering
\caption{Structural evaluation at maximum lag $p=1$ ($N = T - p = 865$).}\label{tab:selected-metrics}
\footnotesize
\setlength{\tabcolsep}{3pt}
\begin{tabular}{@{}lrrrrrr@{}}
\toprule
Method & Max.\ pa. & $d$ & SHD & T-SHD & $BIC$ & $\ell$ \\
\midrule
LASSO & 52 & $7.318\times10^{15}$ & 545 & 543 & $-2.475\times10^{16}$ & -28,717 \\
LAR & 54 & $2.829\times10^{16}$ & 595 & 593 & $-9.565\times10^{16}$ & -28,741 \\
JS & 5 & 637 & 329 & 327 & -32,605 & -30,451 \\
SIMONE & 3 & 191 & 244 & 244 & -32,067 & -31,421 \\
Consensus & 4 & 230 & 247 & 244 & -32,525 & -31,747 \\
Knowledge & 10 & 6,222 & 0 & 0 & -62,784 & -41,746 \\
PC-Stable & 4 & 311 & 281 & 279 & -31,163 & -30,111 \\
GS & 3 & 195 & 242 & 241 & -32,002 & -31,343 \\
IAMB & 4 & 222 & 251 & 250 & -31,880 & -31,129 \\
Fast-IAMB & 3 & 187 & 239 & 238 & -32,015 & -31,383 \\
Inter-IAMB & 4 & 221 & 251 & 250 & -31,597 & -30,850 \\
IAMB-FDR & 3 & 194 & 244 & 244 & -32,086 & -31,430 \\
HC & 14 & 18,217 & 340 & 338 & -90,669 & -29,070 \\
Tabu & 14 & 18,249 & 341 & 339 & -90,772 & -29,066 \\
MMHC & 2 & 179 & 240 & 239 & -31,686 & -31,081 \\
RSMAX2 & 4 & 223 & 255 & 253 & -31,020 & -30,266 \\
H2PC & -- & -- & -- & -- & -- & -- \\
\bottomrule
\end{tabular}
\par\smallskip
\begin{minipage}{\linewidth}\footnotesize\raggedright
Max. pa.: largest parent set. $d$: number of free parameters. SHD: uses the original \texttt{bnlearn::shd} (\cite{bib13}) convention, which compares the corresponding completed partially directed acyclic graphs (CPDAGs). T-SHD: symmetric difference between the actual directed edge sets. A dash denotes an unavailable result.
\end{minipage}
\end{table}

%% file: policy_counts.tex
\begin{table}[htbp]
\centering
\footnotesize
\setlength{\tabcolsep}{3pt}
\caption{Policy-query counts at model maximum lag $p=1$ and exposure lag $\ell=1$.}\label{tab:policy-counts}
\begin{tabular}{@{}llllllll@{}}
\toprule
Method & Computed & Direct & Zero & Covered & Low cov. & No graph & Calc. failed \\
\midrule
LASSO &  36 &  7 &  29 & 0 &  7 &  0 & 0 \\
LAR &  36 &  7 &  29 & 0 &  7 &  0 & 0 \\
JS &  36 &  1 &  35 & 0 &  1 &  0 & 0 \\
SIMONE & 36 & 0 & 36 & 0 & 0 & 0 & 0 \\
Consensus &  36 &  0 &  36 & 0 &  0 &  0 & 0 \\
PC-Stable &  36 &  0 &  36 & 0 &  0 &  0 & 0 \\
GS &  36 &  0 &  36 & 0 &  0 &  0 & 0 \\
IAMB &  36 &  0 &  36 & 0 &  0 &  0 & 0 \\
Fast-IAMB &  36 &  0 &  36 & 0 &  0 &  0 & 0 \\
Inter-IAMB &  36 &  0 &  36 & 0 &  0 &  0 & 0 \\
IAMB-FDR &  36 &  0 &  36 & 0 &  0 &  0 & 0 \\
HC &  36 &  6 &  30 & 4 &  2 &  0 & 0 \\
Tabu &  36 &  6 &  30 & 4 &  2 &  0 & 0 \\
MMHC &  36 &  0 &  36 & 0 &  0 &  0 & 0 \\
RSMAX2 &  36 &  0 &  36 & 0 &  0 &  0 & 0 \\
H2PC &   0 &  0 &   0 & 0 &  0 & 36 & 0 \\
Total & 540 & 27 & 513 & 8 & 19 & 36 & 0 \\
\bottomrule
\end{tabular}
\par\smallskip
\begin{minipage}{\linewidth}
\footnotesize\raggedright
Each listed method has 36 $\Delta_{\ell}$. Covered and low-coverage columns refer only to direct exposure-to-outcome links.
\end{minipage}
\end{table}

%% file: covered_contrasts.tex
\begin{table}[htbp]
\centering
\footnotesize
\setlength{\tabcolsep}{3pt}
\caption{$\Delta_{\ell}$ passing the coverage diagnostic at $p=1$ and $\ell=1$.}\label{tab:covered-contrasts}
\begin{tabular}{@{}llllll@{}}
\toprule
Method & Exposure & Outcome & $100\widehat\Delta$ (pp) & $C_{\rm low}$ & $C_{\rm high}$ \\
\midrule
HC, Tabu & Homeworking & New cases & +1.2239 & 1.0000 & 1.0000 \\
HC, Tabu & Homeworking & New infections & +1.7092 & 1.0000 & 1.0000 \\
HC, Tabu & Grocery/pharmacy & New cases & -0.0442 & 1.0000 & 1.0000 \\
HC, Tabu & Grocery/pharmacy & New infections & -0.1403 & 1.0000 & 1.0000 \\
\bottomrule
\end{tabular}
\par\smallskip
\begin{minipage}{\linewidth}
\footnotesize\raggedright
Multiplication by 100 gives percentage points (pp). $C$ is the fitted-model mass assigned to observed configurations. Coverage passes when unobserved mass is at most 5.0\% under both intervention states. The two methods answer the same question with matching numbers; we print the result once, with both method names.
\end{minipage}
\end{table}

%% file: variables.tex
\begin{longtable}{p{4cm}p{10cm}} 
\caption{The COVID-19 Dataset}\label{tab:colnames} \\
\hline
\textbf{Variable} & \textbf{Description} \\
\hline
\endfirsthead

\hline
\textbf{Variable} & \textbf{Description (continued)} \\
\hline
\endhead

\hline
\endfoot

\hline
\endlastfoot

Date & Date observations were recorded. \\
Excess mortality (EM) & Percentage difference between reported and projected deaths. \\
Schools & Whether schools were open, partially open, or closed. \\
Face masks (FM) & Mask mandates during the pandemic: optional, mandatory, or no mandate. \\
Lockdown severity (LS) & Five categories: severe, moderate, weak, or limited measures/social distancing only. \\
Majority COVID-19 variant (Variant) & The majority COVID-19 variant. \\
Flights 7-day moving average & 7-day moving average of flights. \\
OpenTable restaurant (Restaurant) & OpenTable index on restaurant bookings in London. \\
Google homeworking (Homeworking) & Google index on homeworking in Greater London. \\
Google workplace (Workplace) & Google index on workplace activity in Greater London. \\
Apple walking (Walking) & Apple index on walking activity in London. \\
Google parks (Parks) & Google index on park visits in Greater London. \\
Google retail \& recreation (Retail \& recreation) & Google index on retail and recreation in Greater London. \\
Google grocery \& pharmacy (Grocery \& Pharmacy) & Google index on grocery and pharmacy in Greater London. \\
Google transit stations (Transit) & Google index on transit station activity. \\
TfL Tube (Tube) & TfL index on tube activity. \\
TfL Bus (Bus) & TfL index on bus activity. \\
Citymapper journeys (Journeys) & Citymapper index on journey activity. \\
Season & Winter, autumn, summer, and spring. \\
PCR tests & Number of tests on date. \\
PCR tests capacity & Capacity level on date. \\
Antibody tests & Number of tests on date. \\
Antibody tests capacity & Capacity level on date. \\
Pillar 1 capacity & NHS/UKHSA capacity on date. \\
Pillar 2 capacity & UK Government capacity on date. \\
Pillar 3 capacity & Antibody testing capacity on date. \\
Pillar 4 capacity & Surveillance testing capacity on date. \\
Pillar 1 tests & NHS/UKHSA tests on date. \\
Pillar 2 tests & UK Government tests on date. \\
Pillar 3 tests & Antibody tests on date. \\
Pillar 4 tests & Surveillance tests on date. \\
Tests across all 4 Pillars & Total tests on date. \\
New cases & People testing positive for COVID-19 on date. \\
New infections & New infections on date. \\
Reinfections & New reinfections on date. \\
Hospital admissions & Patients admitted to hospital with COVID-19. \\
Patients in hospital & Patients in hospital with COVID-19. \\
COVID-19 patients in MVBs & Patients in Mechanical Ventilator Beds (MVBs) with COVID-19. \\
Vaccinations (total) & Total vaccines administered on date. \\
Vaccinations (1st dose) & First dose vaccines administered on date. \\
Vaccinations (2nd dose) & Second dose vaccines administered on date. \\
Vaccinations (3rd dose) & Third dose vaccines administered on date. \\
1st dose uptake & Reported first dose uptake. \\
2nd dose uptake & Reported second dose uptake. \\
3rd dose uptake & Reported third dose uptake. \\
COVID-19 deaths on certificate & Daily deaths with COVID-19 on certificate by date of death. \\
\end{longtable}

%% file: missingness.tex
\begin{longtable}{p{8cm}p{2cm}p{2cm}} 
\caption{Summary of Missing Data per Variable} \label{tab:missdata} \\
\hline
\textbf{Variable} & \textbf{Missing Count} & \textbf{Missing Percentage (\%)}\\
\hline
\endfirsthead

\hline
\textbf{Variable} & \textbf{Missing Count} & \textbf{Missing Percentage (\%)} \\
\hline
\endhead

\hline
\endfoot

\hline
\endlastfoot

Date & 0 & 0.00 \\
Excess mortality & 8 & 0.92 \\
Schools & 0 & 0.00 \\
Face masks & 0 & 0.00 \\
Lockdown severity & 0 & 0.00 \\
Majority COVID-19 variant & 5 & 0.58 \\
Flights 7-day moving average & 0 & 0.00 \\
OpenTable restaurant bookings London index & 144 & 16.63 \\
Google homeworking Greater London mobility index & 39 & 4.50 \\
Google workplace Greater London mobility index & 39 & 4.50 \\
Apple walking London mobility index & 247 & 28.52 \\
Google parks Greater London mobility index & 39 & 4.50 \\
Google retail recreation Greater London mobility index & 39 & 4.50 \\
Google grocery pharmacy Greater London mobility index & 39 & 4.50 \\
Google transit stations mobility index & 49 & 5.66 \\
TfL Tube mobility index & 54 & 6.24 \\
TfL Bus mobility index & 54 & 6.24 \\
Citymapper journeys mobility index & 319 & 36.84 \\
Season & 0 & 0.00 \\
PCR tests & 107 & 12.36 \\
PCR tests capacity & 90 & 10.39 \\
Antibody tests & 121 & 13.97 \\
Antibody tests capacity & 70 & 8.08 \\
Pillar 1 NHS and UKHSA capacity & 70 & 8.08 \\
Pillar 2 UK Government capacity & 69 & 7.97 \\
Pillar 3 Antibody capacity & 120 & 13.86 \\
Pillar 4 Surveillance capacity & 70 & 8.08 \\
Pillar 1 NHS and UKHSA tests & 86 & 9.93 \\
Pillar 2 UK Government tests & 86 & 9.93 \\
Pillar 3 Antibody tests & 122 & 14.09 \\
Pillar 4 Surveillance tests & 62 & 7.16 \\
Tests across all 4 Pillars & 86 & 9.93 \\
New cases & 25 & 2.89 \\
New infections & 25 & 2.89 \\
Reinfections & 132 & 15.24 \\
Hospital admissions & 63 & 7.27 \\
Patients in hospital & 64 & 7.39 \\
Patients in MVBs & 85 & 9.82 \\
Vaccinations total & 355 & 40.99 \\
Vaccinations 1st dose & 355 & 40.99 \\
Vaccinations 2nd dose & 355 & 40.99 \\
Vaccinations 3rd dose & 639 & 73.79 \\
First dose uptake & 354 & 40.88 \\
Second dose uptake & 354 & 40.88 \\
Third dose uptake & 617 & 71.25 \\
COVID-19 deaths on certificate & 10 & 1.15 \\
\end{longtable}

%% file: stability_counts.tex
\begin{table}[htbp]
\centering
\small
\caption{Edge recurrence across maximum-lag specifications.}\label{tab:stability-counts}
\begin{tabular}{@{}lrrr@{}}
\toprule
Method & Distinct edges & Meeting rule & Available graphs \\
\midrule
LASSO & 1,466 & 331 & 5 \\
LAR & 3,345 & 684 & 5 \\
JS & 4,640 & 1,460 & 5 \\
SIMONE & 139 & 0 & 5 \\
Consensus & 242 & 90 & 5 \\
PC-Stable & 257 & 136 & 5 \\
GS & 141 & 84 & 5 \\
IAMB & 146 & 82 & 5 \\
Fast-IAMB & 138 & 76 & 5 \\
Inter-IAMB & 151 & 85 & 5 \\
IAMB-FDR & 105 & 72 & 5 \\
HC & 350 & 141 & 5 \\
Tabu & 353 & 144 & 5 \\
MMHC & 92 & 71 & 5 \\
RSMAX2 & 138 & 88 & 5 \\
H2PC & -- & -- & 0 \\
\bottomrule
\end{tabular}
\par\smallskip
\begin{minipage}{\linewidth}\footnotesize\raggedright
The rule requires recurrence $\geq0.8$ in at least 2 successful eligible specifications. Each denominator includes only graphs with maximum lag at least the exposure lag. A dash denotes an unavailable method. The consensus requires 3 votes.
\end{minipage}
\end{table}

%% file: all_direct_contrasts.tex
\begin{table}[htbp]
\centering
\footnotesize
\setlength{\tabcolsep}{3pt}
\caption{All computed direct-link contrasts at $p=1$ and $\ell=1$.}\label{tab:all-policy-contrasts}
\begin{tabular}{@{}lllllll@{}}
\toprule
Method & Exposure & Outcome & $100\widehat\Delta$ (pp) & $C_{\rm low}$ & $C_{\rm high}$ & Coverage \\
\midrule
LASSO & Homeworking & New cases & -3.0617 & 0.1469 & 0.0846 & Low \\
LASSO & Workplace mobility & New cases & -2.2824 & 0.1398 & 0.0954 & Low \\
LASSO & Workplace mobility & New infections & -1.7135 & 0.5912 & 0.5657 & Low \\
LASSO & Workplace mobility & Reinfections & +0.6845 & 0.4466 & 0.4687 & Low \\
LASSO & Grocery/pharmacy & New cases & +3.5385 & 0.0853 & 0.1563 & Low \\
LASSO & Tube & New cases & -4.8997 & 0.1658 & 0.0698 & Low \\
LASSO & Tube & Reinfections & -8.3252 & 0.5116 & 0.3583 & Low \\
LAR & Homeworking & New cases & -16.0090 & 0.6123 & 0.2812 & Low \\
LAR & Workplace mobility & New cases & -6.4031 & 0.5454 & 0.4252 & Low \\
LAR & Workplace mobility & New infections & -4.4799 & 0.7962 & 0.7322 & Low \\
LAR & Workplace mobility & Reinfections & +1.5750 & 0.8400 & 0.8255 & Low \\
LAR & Grocery/pharmacy & New cases & +4.8745 & 0.4595 & 0.5508 & Low \\
LAR & Tube & New cases & -8.0577 & 0.5763 & 0.4183 & Low \\
LAR & Tube & Reinfections & -5.8015 & 0.8340 & 0.8401 & Low \\
JS & Grocery/pharmacy & New infections & +2.4181 & 0.8332 & 0.9127 & Low \\
HC & Homeworking & New cases & +1.2239 & 1.0000 & 1.0000 & Pass \\
HC & Homeworking & New infections & +1.7092 & 1.0000 & 1.0000 & Pass \\
HC & Grocery/pharmacy & New cases & -0.0442 & 1.0000 & 1.0000 & Pass \\
HC & Grocery/pharmacy & New infections & -0.1403 & 1.0000 & 1.0000 & Pass \\
HC & Grocery/pharmacy & Reinfections & +3.0217 & 0.9121 & 0.9461 & Low \\
HC & Tube & Reinfections & -0.0843 & 0.9121 & 0.9656 & Low \\
Tabu & Homeworking & New cases & +1.2239 & 1.0000 & 1.0000 & Pass \\
Tabu & Homeworking & New infections & +1.7092 & 1.0000 & 1.0000 & Pass \\
Tabu & Grocery/pharmacy & New cases & -0.0442 & 1.0000 & 1.0000 & Pass \\
Tabu & Grocery/pharmacy & New infections & -0.1403 & 1.0000 & 1.0000 & Pass \\
Tabu & Grocery/pharmacy & Reinfections & +3.0217 & 0.9121 & 0.9461 & Low \\
Tabu & Tube & Reinfections & -0.0843 & 0.9121 & 0.9656 & Low \\
\bottomrule
\end{tabular}
\par\smallskip
\begin{minipage}{\linewidth}
\footnotesize\raggedright
Multiplication by 100 gives percentage points (pp). $C$ is the fitted-model mass assigned to observed configurations. Coverage passes when unobserved mass is at most 5.0\% under both intervention states.
\end{minipage}
\end{table}